\documentclass[10pt,twocolumn,letterpaper]{article}

\usepackage{cvpr}              

\usepackage{array}
\usepackage{textcomp}
\usepackage{stfloats}
\usepackage{url}
\usepackage{verbatim}
\usepackage{multirow}
\usepackage{color}
\usepackage{balance}

\definecolor{cvprblue}{rgb}{0.21,0.49,0.74}
\usepackage[pagebackref,breaklinks,colorlinks,allcolors=cvprblue]{hyperref}

\def\paperID{*****} 
\def\confName{CVPR}
\def\confYear{2026}

\title{DeltaSeg: Tiered Attention and Deep Delta Learning for Multi-Class Structural Defect Segmentation}

\author{%
Enrique Hernandez Noguera$^{1}$, \quad Md Meftahul Ferdaus$^{1,*}$, \quad Elias Ioup$^{2}$, \quad Mahdi Abdelguerfi$^{1}$ \\[2pt]
$^{1}$University of New Orleans, New Orleans, LA, USA \\
$^{2}$Center for Geospatial Sciences, Naval Research Laboratory, Hancock County, MS, USA \\[2pt]
{\tt\small \{jjlopez3, mferdaus\}@uno.edu, mahdi@cs.uno.edu} \\
{\tt\small \ elias.z.ioup.civ@us.navy.mil} \\[2pt]
$^{*}$Corresponding author%
}

\begin{document}
\maketitle

\begin{abstract}
Automated segmentation of structural defects from visual inspection imagery remains challenging due to the diversity of damage types, extreme class imbalance, and the need for precise boundary delineation. This paper presents DeltaSeg, a U-shaped encoder-decoder architecture with a tiered attention strategy that integrates Squeeze-and-Excitation (SE) channel attention in the encoder, Coordinate Attention at the bottleneck and decoder, and a novel Deep Delta Attention (DDA) mechanism in the skip connections. The encoder uses depthwise separable convolutions with dilated stages to maintain spatial resolution while expanding the receptive field. Atrous Spatial Pyramid Pooling (ASPP) at the bottleneck captures multi-scale context. The DDA module refines skip connections through a dual-path scheme combining a learned delta operator for nuisance feature suppression with spatial attention gates conditioned on decoder signals. Deep supervision through multi-scale auxiliary heads further strengthens gradient flow and encourages semantically meaningful features at intermediate decoder stages. We evaluate DeltaSeg on two datasets: the S2DS dataset (7 classes) and the Culvert-Sewer Defect Dataset (CSDD, 9 classes). Across both benchmarks, DeltaSeg consistently outperforms 12 competing architectures including U-Net, SA-UNet, UNet3+, SegFormer, Swin-UNet, EGE-UNet, FPN, and Mobile-UNETR, demonstrating strong generalization across damage types, imaging conditions, and structural geometries.
\end{abstract}

\section{Introduction}
\label{sec:intro}

Visual inspection of civil infrastructure is essential for assessing structural integrity and planning maintenance. Traditional manual inspection is time-consuming, subjective, and often hazardous~\cite{koch2015review, spencer2019advances}. Computer vision methods, particularly deep learning-based semantic segmentation, offer the potential to automate defect identification at the pixel level, enabling quantitative damage assessment from images captured by drones, robots, or fixed cameras.

Fully convolutional networks (FCNs)~\cite{long2015fcn} and the U-Net architecture~\cite{ronneberger2015unet} established the encoder-decoder paradigm for dense prediction. DeepLab variants~\cite{chen2017deeplabv2, chen2017rethinking, chen2018deeplabv3plus} introduced atrous (dilated) convolution and ASPP modules to capture multi-scale context without sacrificing spatial resolution. Attention mechanisms have further improved segmentation by enabling networks to selectively emphasize informative features. SE-Net~\cite{hu2018squeeze} recalibrates channel responses through global average pooling and gating. CBAM~\cite{woo2018cbam} extends this to spatial attention. Attention U-Net~\cite{oktay2018attention} and its extensions~\cite{schlemper2019attentiongate} use gating signals from the decoder to suppress irrelevant encoder features in skip connections. More recently, Coordinate Attention~\cite{hou2021coordatt} encodes spatial position along two orthogonal directions, jointly capturing channel dependencies and long-range spatial interactions with low computational overhead.

Transformer-based architectures such as SegFormer~\cite{xie2021segformer} and Swin-UNet~\cite{cao2022swinunet} model global context through self-attention, but their computational cost and data requirements can be prohibitive for the relatively small, domain-specific datasets typical in structural health monitoring. Lightweight models like Mobile-UNETR~\cite{hatamizadeh2022unetr} and FPN-based designs~\cite{lin2017fpn, kirillov2019panoptic} reduce parameters but often sacrifice accuracy on fine-grained, multi-class defect boundaries.

Structural defect segmentation poses several specific challenges that general-purpose architectures do not fully address \cite{kuchi2019machine,kuchi2021machine}. First, defect classes such as cracks, spalling, corrosion, efflorescence, and vegetation exhibit large variation in scale, aspect ratio, and texture \cite{panta2023iterlunet,alshawi2025imbalance}. Cracks are thin and elongated; spalling regions are large and irregular; corrosion may appear as scattered patches. Second, class imbalance is severe: background pixels dominate, while some defect classes occupy less than 1\% of image area. Third, practical deployment demands models that generalize across structures (buildings, bridges, culverts) and imaging platforms (UAVs, handheld cameras) without retraining.

This paper addresses these challenges with DeltaSeg, an architecture built on three design principles: (1) use different attention mechanisms where they are most appropriate in the network, (2) refine skip connections to bridge the semantic gap between encoder and decoder, and (3) validate generalization rigorously across two distinct structural deficiency datasets representing different structure types and imaging conditions.

The main contributions are as follows:
\begin{enumerate}
\item A tiered attention strategy that places SE attention in the encoder for channel calibration, Coordinate Attention at the bottleneck and decoder for joint channel-spatial refinement, and a novel Deep Delta Attention module in the skip connections.
\item The Deep Delta Attention (DDA) module, which combines a learned delta operator for nuisance feature suppression with decoder-conditioned spatial gating and multi-scale encoder fusion.
\item A depthwise separable encoder with hybrid pooling and dilation that balances computational efficiency with receptive field expansion.
\item Comprehensive evaluation on two datasets spanning 7 to 9 classes, covering buildings and culverts, with comparisons against 12 architectures.
\end{enumerate}

\section{Related Work}
\label{sec:related}

\subsection{Encoder-Decoder Segmentation}
U-Net~\cite{ronneberger2015unet} introduced symmetric skip connections between encoder and decoder, enabling precise spatial recovery. DeepLabv3+~\cite{chen2018deeplabv3plus} combined an ASPP module with an encoder-decoder structure using dilated convolutions, achieving strong results on natural image benchmarks. Depthwise separable convolutions, introduced in MobileNets~\cite{howard2017mobilenets} and formalized in Xception~\cite{chollet2017xception}, factorize standard convolutions into spatial and channel components, reducing computation by an order of magnitude with minimal accuracy loss.

\subsection{Attention Mechanisms in Segmentation}
SE-Net~\cite{hu2018squeeze} computes channel-wise attention through global pooling followed by a two-layer bottleneck with sigmoid gating. CBAM~\cite{woo2018cbam} adds a spatial attention branch using max and average pooling. ECA-Net~\cite{wang2020eca} simplifies channel attention to a 1D convolution. Coordinate Attention~\cite{hou2021coordatt} decomposes the spatial attention problem along the horizontal and vertical axes, encoding positional information while preserving channel dependencies. This directional decomposition is effective for elongated structures.

Attention gates~\cite{oktay2018attention, schlemper2019attentiongate} use decoder features as gating signals to highlight salient encoder regions in skip connections. PSPNet~\cite{zhao2017pspnet} aggregates multi-scale context through pyramid pooling. These methods apply a single attention type uniformly across the network. In contrast, our approach assigns different attention mechanisms to different network stages based on the information needs at each location.

\subsection{Structural Defect Segmentation}
Early work on crack detection used edge filters and morphological operations~\cite{koch2015review}. Learning-based approaches, including FCN~\cite{long2015fcn} and U-Net variants, significantly improved performance. Attention-augmented architectures such as SA-UNet~\cite{sun2022saunet}, originally proposed for medical image segmentation, have since been adopted as competitive baselines in structural defect benchmarks owing to their spatial attention mechanism in the encoder. The S2DS dataset~\cite{hoskere2020s2ds} established a multi-class benchmark for structural damage with seven categories. The Culvert-Sewer Defect Dataset (CSDD)~\cite{csdd2024} provides annotated culvert inspection frames with nine defect classes following the NASSCO PACP standard.

Prior work in this domain typically applies a single attention mechanism throughout the network or uses simple concatenation-based skip connections. The DDA module proposed here processes skip connections through dual parallel attention paths conditioned on decoder signals, directly addressing the semantic gap between encoder features and decoder reconstruction targets.

\section{Problem Formulation}
\label{sec:problem}

Let $\mathcal{X} = \mathbb{R}^{H \times W \times 3}$ denote the space of RGB input images and $\mathcal{Y} = \{0, 1, \ldots, C-1\}^{H \times W}$ the space of pixel-wise label maps, where $H$, $W$ are the spatial dimensions and $C$ is the number of semantic classes. The goal of semantic segmentation is to learn a mapping $f_\theta: \mathcal{X} \rightarrow \mathcal{P}$, parameterized by $\theta$, that assigns each pixel a probability distribution over $C$ classes, where $\mathcal{P} = [0,1]^{H \times W \times C}$ with $\sum_{c=0}^{C-1} f_\theta(\mathbf{x})_{i,j,c} = 1$ for all spatial locations $(i,j)$.

Given a training set $\mathcal{D} = \{(\mathbf{x}_n, \mathbf{y}_n)\}_{n=1}^{N}$ of image--label pairs, we seek parameters $\theta^*$ that minimize the empirical risk:
\begin{equation}
\theta^* = \arg\min_\theta \frac{1}{N} \sum_{n=1}^{N} \mathcal{L}\bigl(f_\theta(\mathbf{x}_n),\, \mathbf{y}_n\bigr),
\label{eq:objective}
\end{equation}
where $\mathcal{L}$ is a composite loss combining pixel-wise cross-entropy and a region-based Dice term (detailed in \cref{sec:method}).

Structural defect segmentation introduces several challenges beyond those encountered in natural-scene parsing. First, the \textit{extreme class imbalance}: background pixels dominate each image, while defect classes such as cracks or efflorescence occupy thin, irregularly shaped regions. Second, \textit{fine-grained boundary delineation} is critical---the practical value of a defect map depends on accurately localizing defect extents rather than merely detecting their presence. Third, defect classes exhibit \textit{high intra-class variability} in appearance due to differences in material, lighting, weathering, and viewpoint, yet share \textit{subtle inter-class boundaries} (e.g., spalling versus corrosion on the same surface).

These properties impose specific requirements on the segmentation architecture. The encoder must capture multi-scale context to handle defects ranging from hairline cracks to large spalled regions, while the decoder must recover spatial detail lost during downsampling. Critically, the skip connections should selectively propagate boundary-relevant features rather than naively forwarding all encoder activations, which we formalize through a learned attention-gated skip mechanism in \cref{sec:method}.

\section{Methodology}
\label{sec:method}

\begin{figure}[!t]
\centering
\includegraphics[width=\columnwidth]{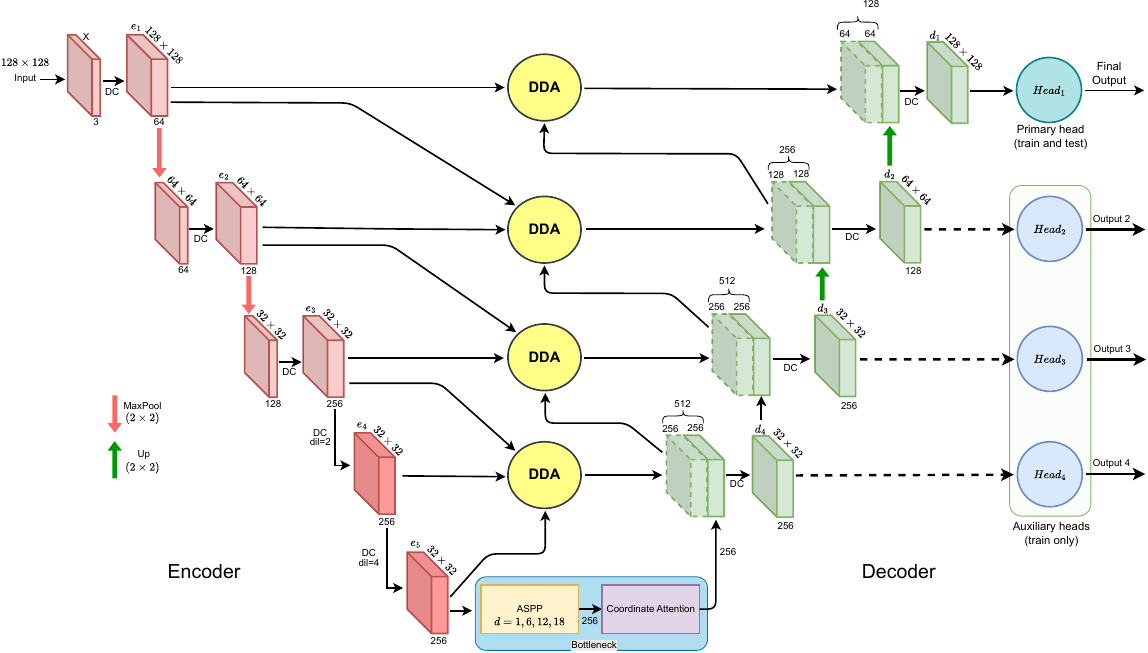}
\caption{DeltaSeg architecture. The encoder uses DSC with SE attention, switching to dilated convolutions in stages 4--5. ASPP captures multi-scale context at the bottleneck. DDA modules refine skip connections via dual-path attention. The decoder upsamples and fuses DDA-refined features with Coordinate Attention. DC = DoubleConv (two DSC--BN--ReLU units). Auxiliary heads provide deep supervision during training.}
\label{fig:architecture}
\end{figure}

We present DeltaSeg, a U-Net-based architecture that realizes the mapping $f_\theta$ defined in \cref{sec:problem} through a tiered attention design: channel attention (SE) in the encoder, positional attention (Coordinate Attention) in the bottleneck and decoder, and a novel Deep Delta Attention (DDA) mechanism in the skip connections. \cref{fig:architecture} provides an overview. The network accepts images $\mathbf{x} \in \mathbb{R}^{H \times W \times 3}$ and produces dense predictions over $C$ classes; the spatial resolution $H{\times}W$ is dataset-dependent ($256{\times}256$ for S2DS, $128{\times}128$ for CSDD), as detailed in \cref{sec:experiments}.

\begin{figure}[t]
    \centering
    \includegraphics[width=\linewidth]{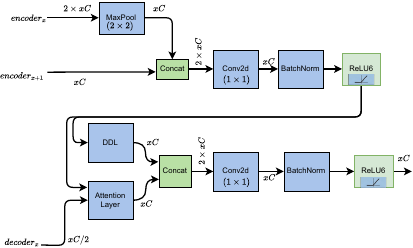}
    \caption{Internal structure of the Deep Delta Attention (DDA) module. Given encoder and decoder features, DDA computes a refined skip feature via dual-path attention---Deep Delta Learning (above) and attention gating (below)---to emphasize boundary-relevant differences before decoder fusion.}
    \label{fig:dda}
\end{figure}

\subsection{Depthwise Separable Convolution Block}
\label{sec:dsc}
All convolutional stages employ \emph{depthwise separable convolutions} (DSC), which factorize a standard convolution into a depthwise $K{\times}K$ spatial filter followed by a pointwise $1{\times}1$ cross-channel projection. Each operation is followed by batch normalization (BatchNorm or BN)~\cite{ioffe2015batchnorm} and ReLU6 activation. A \textit{DoubleConv} block stacks two such DSC units in sequence, reducing the parameter count per block from $\mathcal{O}(K^2 C_\text{in} C_\text{out})$ to $\mathcal{O}(K^2 C_\text{in} + C_\text{in} C_\text{out})$---an approximate $9\times$ reduction for $K{=}3$ at typical channel widths. Each DoubleConv block is followed by a stage-specific attention module, described below.

\subsection{Encoder}
\label{sec:encoder}

The encoder comprises five stages that produce a hierarchy of feature tensors $\{\mathbf{F}^{(l)}_\text{enc}\}_{l=1}^{5}$. Stages 1--3 use standard depthwise separable convolutions ($dilation (d) {=}1$) interleaved with $2{\times}2$ max pooling, yielding feature maps of spatial resolutions $128^2$, $64^2$, and $32^2$ (CSDD) with channel dimensions 64, 128, and 256, respectively. Stages 4--5 replace pooling with dilated convolutions at rates $d{=}2$ and $d{=}4$, maintaining $32^2$ spatial resolution while expanding the effective receptive field to $5{\times}5$ and $9{\times}9$, respectively. This hybrid design addresses the requirement identified in \cref{sec:problem}: capturing multi-scale context for defects of varying spatial extent while preserving the fine spatial detail necessary for boundary delineation.
Maintaining $32^2$ resolution at these stages is deliberate: further spatial downsampling would reduce feature maps to $16^2$ or $8^2$, at which point dilation rates $d{>}1$ become ineffective---the dilated kernel spans the entire feature map or samples from only a handful of distinct locations, collapsing the multi-scale receptive field expansion to a degenerate operation.

\paragraph{Squeeze-and-Excitation (SE) attention.} Each encoder DoubleConv DSC unit is followed by an SE module~\cite{hu2018squeeze} that performs channel-wise recalibration. Given the input feature tensor $\mathbf{F} \in \mathbb{R}^{C \times H \times W}$, SE computes:
\begin{equation}
\mathbf{z} = \text{GAP}(\mathbf{F}) \in \mathbb{R}^{C}, \qquad
\mathbf{s} = \sigma\!\bigl(\mathbf{W}_2 \, \text{ReLU}(\mathbf{W}_1 \mathbf{z})\bigr) \in \mathbb{R}^{C},
\label{eq:se}
\end{equation}
where $\text{GAP}(\cdot)$ denotes global average pooling, which computes the spatial mean $\frac{1}{HW}\sum_{i,j}\mathbf{F}(c,i,j)$ independently per channel to produce a $C$-dimensional descriptor; $\mathbf{W}_1 \in \mathbb{R}^{(C/r) \times C}$, $\mathbf{W}_2 \in \mathbb{R}^{C \times (C/r)}$, and $r{=}8$ is the reduction ratio (bottleneck dimension clamped to $\max(C/r, 8)$). The recalibrated output is $\hat{\mathbf{F}}(c,:,:) = \mathbf{s}(c) \cdot \mathbf{F}(c,:,:)$. We adopt SE in the encoder because feature extraction at this stage primarily requires \emph{channel selectivity}---identifying which feature types (edges, textures, color distributions) are task-relevant---while spatial localization is deferred to the decoder path.

\subsection{ASPP Bottleneck with Coordinate Attention}
\label{sec:bottleneck}
The encoder output $\mathbf{F}^{(5)}_\text{enc} \in \mathbb{R}^{256 \times 32 \times 32}$ is processed by an Atrous Spatial Pyramid Pooling (ASPP) module~\cite{chen2017rethinking} consisting of five parallel branches: a $1{\times}1$ convolution, three $3{\times}3$ atrous convolutions at dilation rates $d \in \{6, 12, 18\}$ with effective receptive fields of approximately 13, 25, and 37 pixels, and a global average pooling branch followed by $1{\times}1$ projection and bilinear upsampling; each branch is followed by Batch Normalization and ReLU, outputting 256 channels. The concatenated 1280-channel representation is projected back to 256 channels via $1{\times}1$ convolution with 50\% dropout.

\paragraph{Coordinate Attention (CA).} Unlike SE, which discards all spatial structure, CA~\cite{hou2021coordatt} preserves positional information by decomposing global attention into two directional components. Given $\mathbf{F} \in \mathbb{R}^{C \times H \times W}$, horizontal and vertical encodings are computed via directional average pooling:
\begin{equation}
\mathbf{z}^h(c, i) = \frac{1}{W}\textstyle\sum_{j=1}^{W} \mathbf{F}(c,i,j),
\end{equation}
\begin{equation}
\mathbf{z}^w(c, j) = \frac{1}{H}\textstyle\sum_{i=1}^{H} \mathbf{F}(c,i,j).
\label{eq:ca_pool}
\end{equation}
The two vectors are concatenated to form a joint representation of shape $C \times (H{+}W)$, compressed to $C/r$ channels via a shared $1{\times}1$ convolution with batch normalization and ReLU6, then split and independently projected back to $C$ channels through separate $1{\times}1$ convolutions followed by sigmoid:
\begin{equation}
\mathbf{a}^h = \sigma\!\bigl(\text{Conv}_{1\times1}(\mathbf{f}^h)\bigr) \in \mathbb{R}^{C \times H \times 1},
\end{equation}
\begin{equation}
\mathbf{a}^w = \sigma\!\bigl(\text{Conv}_{1\times1}(\mathbf{f}^w)\bigr) \in \mathbb{R}^{C \times 1 \times W}.
\label{eq:ca_attn}
\end{equation}
The output is obtained by element-wise scaling: $\hat{\mathbf{F}}(c,i,j) = \mathbf{F}(c,i,j) \cdot \mathbf{a}^h(c,i) \cdot \mathbf{a}^w(c,j)$. The shared bottleneck forces height and width representations to interact, enabling the network to learn cross-directional dependencies. CA is applied at the bottleneck, after the ASPP module, to ensure the decoder receives features encoding both channel relevance and spatial focus---a critical property for the transition from feature extraction to spatial reconstruction.

\subsection{Deep Delta Attention (DDA) Skip Connections}
\label{sec:dda}

Standard U-Net skip connections forward encoder features to the decoder via concatenation, ignoring the semantic gap between low-level encoder representations and high-level decoder features. DDA bridges this gap through two complementary mechanisms---\emph{Deep Delta Learning} (DDL) and \emph{attention gating}---operating in parallel on multi-scale encoder features (\cref{fig:dda}).

\paragraph{Input specification.} Each DDA module at level $l$ receives: (i) the encoder feature $\mathbf{F}^{(l)}_\text{enc} \in \mathbb{R}^{C \times H \times W}$; (ii) the adjacent shallower encoder feature $\mathbf{F}^{(l+1)}_\text{enc} \in \mathbb{R}^{C' \times 2H \times 2W}$, providing multi-scale context; and (iii) the decoder feature $\mathbf{F}^{(l)}_\text{dec} \in \mathbb{R}^{C/2 \times H' \times W'}$ from the preceding decoder stage, serving as a gating signal. At the shallowest level ($l{=}1$), $\mathbf{F}^{(l+1)}_\text{enc}$ is absent and multi-scale fusion is skipped.

\paragraph{Multi-scale fusion.} When $\mathbf{F}^{(l+1)}_\text{enc}$ is available, it is spatially aligned to $\mathbf{F}^{(l)}_\text{enc}$ via $2{\times}2$ max pooling, concatenated along the channel axis, and projected back to $C$ channels:
\begin{equation}
\mathbf{F}_\text{fused} = \text{ReLU6}\!\bigl(\text{BN}\bigl(\text{Conv}_{1\times1}^{2C \to C}([\mathbf{F}^{(l)}_\text{enc};\; \text{Pool}(\mathbf{F}^{(l+1)}_\text{enc})])\bigr)\bigr).
\label{eq:dda_fuse}
\end{equation}
The $1{\times}1$ projection reduces the $2C$-channel concatenation back to $C$ channels, followed by BN and ReLU6 to normalize the merged representation and introduce the non-linearity needed to learn cross-scale interactions.

\paragraph{Deep Delta Learning (DDL) path.} The DDL path learns to suppress a nuisance direction in channel space through two independent branches. The \emph{direction branch} computes a unit vector $\mathbf{k} \in \mathbb{R}^C$ identifying the nuisance axis, while the \emph{strength branch} produces a scalar $\beta \in [0, 2]$ controlling suppression magnitude:
\begin{equation}
\mathbf{k} = \frac{\text{MLP}_k(\text{GAP}(\mathbf{F}_\text{fused}))}{\|\text{MLP}_k(\text{GAP}(\mathbf{F}_\text{fused}))\|_2},
\end{equation}
\begin{equation}
\beta = 2 \cdot \sigma\!\bigl(\text{MLP}_\beta(\text{GAP}(\mathbf{F}_\text{fused}))\bigr).
\label{eq:ddl}
\end{equation}
where each MLP consists of $1{\times}1$ conv--ReLU--$1{\times}1$ conv applied to the global average pooled (GAP) input. The refined feature is obtained by projecting out the nuisance component:
\begin{equation}
\mathbf{F}_\text{delta} = \mathbf{F}_\text{fused} - \beta \cdot (\mathbf{F}_\text{fused}^\top \mathbf{k})\, \mathbf{k}.
\label{eq:delta_apply}
\end{equation}
Geometrically, $\mathbf{k}$ identifies the axis in channel space carrying uninformative patterns (e.g., dominant background texture), and $\beta$ modulates the degree of suppression. The two-branch design allows direction and magnitude to be estimated independently.

\paragraph{Attention gate path.} The attention gate leverages the decoder signal $\mathbf{g} = \mathbf{F}^{(l)}_\text{dec}$ to determine which spatial locations in the encoder features are relevant to the reconstruction objective. Both $\mathbf{F}_\text{fused}$ and $\mathbf{g}$ are projected to a shared $C$-dimensional space via $3{\times}3$ convolutions (with stride-2 on the encoder side when resolutions differ), summed, and passed through PReLU and a $1{\times}1$ convolution to produce a spatial attention map:
\begin{equation}
\boldsymbol{\alpha} = \sigma\!\bigl(\text{Conv}_{1\times1}^{C \to 1}(\text{PReLU}(\mathbf{g}' + \mathbf{F}'_\text{fused}))\bigr) \in [0,1]^{H_d \times W_d}.
\label{eq:ag}
\end{equation}
When the encoder resolution exceeds the decoder resolution, $\boldsymbol{\alpha}$ is upsampled via transposed convolution before element-wise multiplication: $\mathbf{F}_\text{gate} = \text{Up}(\boldsymbol{\alpha}) \odot \mathbf{F}_\text{fused}$.

\paragraph{Path combination.} The DDL and attention gate outputs are concatenated and projected to $C$ channels:
\begin{equation}
\mathbf{F}_\text{DDA} = \text{ReLU6}\!\bigl(\text{BN}\bigl(\text{Conv}_{1\times1}^{2C \to C}([\mathbf{F}_\text{delta};\; \mathbf{F}_\text{gate}])\bigr)\bigr).
\label{eq:dda_combine}
\end{equation}
This dual-path design directly addresses the challenges outlined in \cref{sec:problem}: the DDL path suppresses class-imbalanced background features, while the attention gate selectively propagates boundary-relevant spatial information conditioned on decoder context.

\subsection{Decoder}
\label{sec:decoder}
The decoder consists of four stages that progressively reconstruct spatial resolution. At each stage $l$, the DDA-refined skip $\mathbf{F}^{(l)}_\text{DDA}$ is concatenated with the upsampled decoder feature from stage $l{+}1$ and processed by a DoubleConv block with CA instead of SE. Stages dec4--dec3 operate at $32{\times}32$; dec2--dec1 upsample to $64{\times}64$ and $128{\times}128$ via transposed convolution (channel progression $256 \to 256 \to 128 \to 64$). CA is used in the decoder because spatial reconstruction requires positional awareness that SE's spatially-collapsed representation cannot provide.

\subsection{Deep Supervision and Loss Function}
\label{sec:loss}
DeltaSeg employs four $1{\times}1$ prediction heads attached to decoder stages dec1--dec4~\cite{lee2019deeply}. The primary head (dec1, full resolution) produces the inference output, while auxiliary heads (dec2--dec4) are active only during training, with their predictions bilinearly upsampled to full resolution.

The loss in \cref{eq:objective} is instantiated as a weighted composite of class-weighted cross-entropy (CE), Dice loss~\cite{sudre2017generalised}, and focal loss~\cite{lin2017focal}:
\begin{equation}
\mathcal{L}_\text{primary} = 0.5\,\mathcal{L}_\text{CE} + 0.3\,\mathcal{L}_\text{Dice} + 0.2\,\mathcal{L}_\text{Focal},
\label{eq:loss}
\end{equation}
where CE uses inverse-frequency class weights, Dice optimizes region overlap to mitigate class imbalance, and focal loss concentrates gradient signal on hard boundary pixels. The total objective augments the primary loss with weighted auxiliary terms:
\begin{equation}
\mathcal{L}_\text{total} = \frac{1}{\sum_k \lambda_k} \sum_{k=1}^{4} \lambda_k \, \mathcal{L}^{(k)},
\label{eq:total_loss}
\end{equation}
where $\lambda_1{=}1.0$, $\lambda_2{=}0.8$, $\lambda_3{=}0.6$, $\lambda_4{=}0.4$ for dec1 through dec4, respectively, and $\sum_k \lambda_k = 2.8$. The decreasing weights assign the highest importance to the full-resolution primary head while still propagating gradient signal through intermediate decoder stages, encouraging semantically meaningful features at all levels.

\section{Experimental Setup}
\label{sec:experiments}

\subsection{Datasets}
\label{sec:datasets}

\subsubsection{Augmented S2DS}
The Structural Defects Dataset (S2DS)~\cite{hoskere2020s2ds} comprises 743 images of concrete surfaces annotated into seven classes: background, crack, spalling, corrosion, efflorescence, vegetation, and control point (563/87/93 train/val/test split). S2DS exhibits significant class imbalance and varied imaging conditions. We apply an augmentation pipeline (flips, affine warping, rotations, brightness/contrast adjustment, Gaussian blur, noise injection, random resized cropping) that expands the training set to 3{,}378 images.

\subsubsection{CSDD (Culvert-Sewer Defect Dataset)}
The Culvert-Sewer Defect Dataset (CSDD)~\cite{csdd2024} consists of 6{,}300 frames from 580 inspection videos (70/15/15\% train/val/test), annotated into nine NASSCO PACP classes: background, crack, root intrusion, hole, joint issue, deformation, fracture, encrustation, and loose gasket. Its cylindrical culvert geometry differs fundamentally from S2DS facades, providing a stringent test of generalization.

\subsection{Competing Architectures}
We compare DeltaSeg against 12 architectures spanning four categories: (1) standard encoder-decoders (U-Net~\cite{ronneberger2015unet}, UNet3+~\cite{huang2020unet3p}); (2) attention-based models (SA-UNet~\cite{sun2022saunet}, EGE-UNet~\cite{ruan2023ege}); (3) transformer-based architectures (SegFormer~\cite{xie2021segformer}, Swin-UNet~\cite{cao2022swinunet}, Hi-ViT-UNet~\cite{zhang2023hivit}, VM-UNet~\cite{ruan2024vmunet}); and (4) lightweight models (FPN~\cite{lin2017fpn}, Mobile-UNETR~\cite{hatamizadeh2022unetr}, EfficientViT~\cite{cai2023efficientvit}, EfficientSegNet~\cite{zhang2025effsegnet}). All models are trained from scratch on both datasets using identical augmentation, loss, and training schedule. Where published architectures were substantially smaller, channel widths were scaled to a comparable parameter range (e.g., SA-UNet from sub-1M to 7.86M) to ensure performance differences reflect design choices rather than capacity disparities.

\subsection{Training Details}
All models are trained for 100 epochs with AdamW~\cite{loshchilov2019adamw} (lr $= 10^{-3}$, weight decay $10^{-5}$, cosine annealing). Input images are resized to $256{\times}256$ for S2DS and $128{\times}128$ for CSDD with batch size 16.

\subsection{Evaluation Metrics}
We report defect-only mean Intersection over Union (mIoU) and mean Dice coefficient (equivalently F1-score). Defect mIoU---averaging IoU across defect classes only, excluding background---is the primary metric. Since all architectures achieve background IoU above 90\% on both datasets, including it uniformly inflates scores and compresses inter-model differences.

\section{Results and Discussion}
\label{sec:results}

\subsection{Results on Augmented S2DS}

\cref{tab:s2ds} reports defect-only mIoU for all architectures on the augmented S2DS dataset. DeltaSeg achieves the highest defect mIoU of 70.46\% and the highest F1-score of 83.99\%, outperforming the next best method (SA-UNet, 70.27\%) by a small but consistent margin. Among the competing architectures, SA-UNet and UNet3+ perform strongly, while transformer-based models (SegFormer, Swin-UNet) and the standard U-Net lag considerably, with Swin-UNet achieving only 18.13\% defect mIoU, reflecting near-zero performance on the vegetation and control point classes rather than a true overall segmentation capability.

\begin{table}[!t]
\centering
\caption{Overall performance comparison on Augmented S2DS (7 classes). Best results in \textbf{bold}, second best \underline{underlined}.}
\label{tab:s2ds}
\resizebox{\columnwidth}{!}{%
\begin{tabular}{@{}lccc@{}}
\toprule
\textbf{Method} & \textbf{Params (M)} & \textbf{Def. mIoU (\%)} & \textbf{F1 (\%)} \\
\midrule
Swin-UNet~\cite{cao2022swinunet} & 14.53 & 18.13 & 33.59 \\
U-Net~\cite{ronneberger2015unet} & 31.04 & 31.54 & 49.77 \\
VM-UNet~\cite{ruan2024vmunet} & 29.08 & 52.47 & 68.36 \\
EfficientViT~\cite{cai2023efficientvit} & 0.27 & 53.45 & 69.34 \\
EfficientSegNet~\cite{zhang2025effsegnet} & 0.29 & 53.64 & 69.28 \\
SegFormer~\cite{xie2021segformer} & 2.67 & 58.20 & 74.56 \\
Hi-ViT-UNet~\cite{zhang2023hivit} & 14.77 & 59.60 & 76.03 \\
FPN~\cite{lin2017fpn} & 21.18 & 62.88 & 78.53 \\
EGE-UNet~\cite{ruan2023ege} & 2.83 & 63.23 & 78.82 \\
Mob.-UNETR~\cite{hatamizadeh2022unetr} & 12.71 & 65.77 & 79.93 \\
UNet3+~\cite{huang2020unet3p} & 25.59 & 68.79 & 82.78 \\
SA-UNet~\cite{sun2022saunet} & 7.86 & \underline{70.27} & \underline{83.45} \\
\midrule
\textbf{DeltaSeg (Ours)} & \textbf{7.14} & \textbf{70.46} & \textbf{83.99} \\
\bottomrule
\end{tabular}
}
\end{table}

\cref{tab:s2ds_class} presents the per-class IoU breakdown. Several architectures fail entirely on vegetation and control point---the two rarest categories---underscoring the difficulty of small-area defect segmentation. DeltaSeg achieves the best vegetation IoU (63.22\%) and second-best control point (42.45\%), demonstrating the benefit of Coordinate Attention and DDA skip connections for preserving fine-grained features relevant to these challenging classes.

\begin{table}[!t]
\centering
\caption{Per-class IoU (\%) on Augmented S2DS. BG = Background, Cra = Crack, Spa = Spalling, Cor = Corrosion, Eff = Efflorescence, Veg = Vegetation, CP = Control Point. Best in \textbf{bold}, second best \underline{underlined}.}
\label{tab:s2ds_class}
\resizebox{\columnwidth}{!}{%
\begin{tabular}{@{}lccccccc|c@{}}
\toprule
\textbf{Method} & \textbf{BG} & \textbf{Cra} & \textbf{Spa} & \textbf{Cor} & \textbf{Eff} & \textbf{Veg} & \textbf{CP} & \textbf{Def. mIoU} \\
\midrule
Swin-UNet~\cite{cao2022swinunet} & 92.22 & 64.50 & 36.09 & 71.20 & 86.30 & 0.00 & 0.00 & 18.13 \\
U-Net~\cite{ronneberger2015unet} & 93.28 & 30.64 & 38.47 & 75.74 & 42.66 & 1.73 & 0.00 & 31.54 \\
VM-UNet~\cite{ruan2024vmunet} & 95.01 & 53.13 & 54.86 & 76.13 & 86.03 & 44.50 & 0.14 & 52.47 \\
Eff.ViT~\cite{cai2023efficientvit} & 94.06 & 53.63 & 62.39 & 73.72 & 78.30 & 52.59 & 0.07 & 53.45 \\
Eff.SegNet~\cite{zhang2025effsegnet} & 95.05 & 50.79 & 54.37 & 73.82 & 86.81 & 56.05 & 0.00 & 53.64 \\
SegFormer~\cite{xie2021segformer} & 94.74 & 52.17 & 5.78 & 76.55 & 85.56 & \underline{60.55} & 16.61 & 58.20 \\
Hi-ViT-UNet~\cite{zhang2023hivit} & 95.08 & 59.75 & 63.57 & 73.42 & 90.37 & 45.37 & 25.12 & 59.60 \\
FPN~\cite{lin2017fpn} & 95.86 & 69.73 & 62.94 & 74.01 & 92.76 & 46.81 & 31.02 & 62.88 \\
EGE-UNet~\cite{ruan2023ege} & 96.31 & 68.48 & 63.75 & 75.30 & 93.81 & 45.04 & 33.00 & 63.23 \\
Mob.-UNETR~\cite{hatamizadeh2022unetr} & \underline{96.94} & 78.28 & 67.19 & \textbf{79.40} & 91.42 & 54.83 & 23.49 & 65.77 \\
UNet3+~\cite{huang2020unet3p} & 96.93 & \underline{80.21} & 65.89 & 75.36 & \textbf{96.66} & 50.49 & \textbf{44.13} & 68.79 \\
SA-UNet~\cite{sun2022saunet} & \textbf{97.31} & \textbf{83.80} & \textbf{72.31} & \underline{76.69} & \underline{96.57} & 54.45 & 37.76 & \underline{70.27} \\
\midrule
\textbf{DeltaSeg} & 96.94 & 74.33 & \underline{71.66} & 74.69 & 96.40 & \textbf{63.22} & \underline{42.45} & \textbf{70.46} \\
\bottomrule
\end{tabular}
}
\end{table}

\begin{figure}[!t]
\centering
\includegraphics[width=\columnwidth]{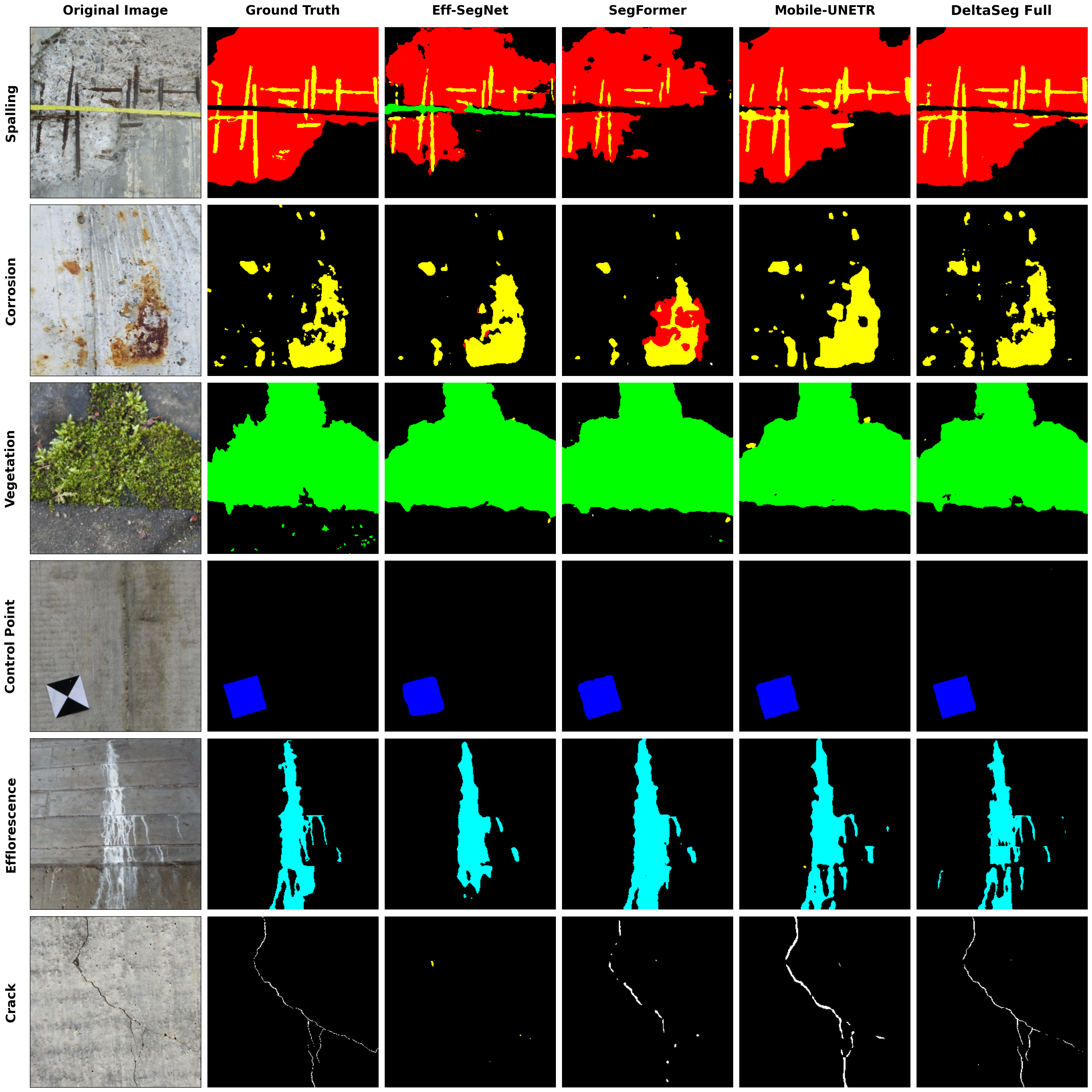}
\caption{Qualitative comparison on S2DS. Rows: defect classes. Columns: image, ground truth, EfficientSegNet, SegFormer, Mobile-UNETR, DeltaSeg (Ours).}
\label{fig:model_comparison}
\end{figure}

\cref{fig:model_comparison} provides a qualitative comparison on six S2DS defect classes. DeltaSeg consistently produces the most spatially coherent masks: it preserves spalling boundaries most faithfully, avoids the class confusion exhibited by SegFormer on corrosion, and recovers the full crack topology---the most demanding class due to its thin, branching structure---more completely than EfficientSegNet and SegFormer.

\subsection{Results on CSDD}

\cref{tab:csdd} presents the CSDD results. DeltaSeg achieves the highest defect mIoU of 76.75\% and F1 of 87.61\%, surpassing UNet3+ (74.22\%) and SA-UNet (74.21\%). The larger margin over external competitors on CSDD than on S2DS suggests that the DDA module and tiered attention are particularly beneficial for the complex cylindrical geometry of culvert imagery.

\begin{table}[!t]
\centering
\caption{Overall performance comparison on CSDD (9 classes). Best results in \textbf{bold}, second best \underline{underlined}.}
\label{tab:csdd}
\resizebox{\columnwidth}{!}{%
\begin{tabular}{@{}lccc@{}}
\toprule
\textbf{Method} & \textbf{Params (M)} & \textbf{Def. mIoU (\%)} & \textbf{F1 (\%)} \\
\midrule
VM-UNet~\cite{ruan2024vmunet} & 29.08 & 63.28 & 78.63 \\
SegFormer~\cite{xie2021segformer} & 2.67 & 66.70 & 80.88 \\
EfficientViT~\cite{cai2023efficientvit} & 0.22 & 67.54 & 81.81 \\
EfficientSegNet~\cite{zhang2025effsegnet} & 0.29 & 67.98 & 82.12 \\
Hi-ViT-UNet~\cite{zhang2023hivit} & 14.58 & 69.28 & 83.10 \\
Swin-UNet~\cite{cao2022swinunet} & 14.52 & 71.09 & 84.19 \\
EGE-UNet~\cite{ruan2023ege} & 2.83 & 71.19 & 84.18 \\
FPN~\cite{lin2017fpn} & 21.20 & 72.17 & 84.90 \\
U-Net~\cite{ronneberger2015unet} & 31.04 & 72.62 & 85.22 \\
Mob.-UNETR~\cite{hatamizadeh2022unetr} & 12.71 & 72.90 & 85.48 \\
SA-UNet~\cite{sun2022saunet} & 7.86 & 74.21 & \underline{86.21} \\
UNet3+~\cite{huang2020unet3p} & 25.59 & \underline{74.22} & 86.14 \\
\midrule
\textbf{DeltaSeg (Ours)} & \textbf{7.14} & \textbf{76.75} & \textbf{87.61} \\
\bottomrule
\end{tabular}
}
\end{table}

\cref{tab:csdd_class} presents the per-class IoU on CSDD. DeltaSeg attains top scores on six of nine classes and leads by clear margins on geometrically distinctive defects such as root intrusion and deformation, demonstrating robust performance on defects unique to culvert structures.

\begin{table}[!t]
\centering
\caption{Per-class IoU (\%) on CSDD. BG = Background, Cra = Crack, Hol = Hole, Roo = Root Intrusion, Def = Deformation, Fra = Fracture, Ero = Erosion, Joi = Joints, Gas = Loose Gasket. Best in \textbf{bold}, second best \underline{underlined}.}
\label{tab:csdd_class}
\resizebox{\columnwidth}{!}{%
\begin{tabular}{@{}l*{9}{c}|c@{}}
\toprule
\textbf{Method} & \rotatebox{70}{\textbf{BG}} & \rotatebox{70}{\textbf{Cra}} & \rotatebox{70}{\textbf{Hol}} & \rotatebox{70}{\textbf{Roo}} & \rotatebox{70}{\textbf{Def}} & \rotatebox{70}{\textbf{Fra}} & \rotatebox{70}{\textbf{Ero}} & \rotatebox{70}{\textbf{Joi}} & \rotatebox{70}{\textbf{Gas}} & \rotatebox{70}{\textbf{Def. mIoU}} \\
\midrule
VM-UNet~\cite{ruan2024vmunet} & 97.0 & 43.8 & 86.9 & 81.1 & 73.5 & 50.9 & 36.3 & 65.5 & 68.4 & 63.28 \\
SegFormer~\cite{xie2021segformer} & 97.0 & 45.5 & 94.7 & 83.3 & 73.5 & 52.1 & 41.8 & 69.9 & 72.9 & 66.70 \\
Eff.ViT~\cite{cai2023efficientvit} & 96.7 & 43.3 & 92.2 & 82.0 & 71.3 & 53.6 & 64.3 & 65.7 & 67.9 & 67.54 \\
Eff.SegNet~\cite{zhang2025effsegnet} & 97.4 & 45.9 & 95.0 & 83.7 & 60.7 & 55.8 & 66.2 & 72.9 & 63.5 & 67.98 \\
Hi-ViT-UNet~\cite{zhang2023hivit} & 97.4 & 53.8 & 93.9 & 85.9 & 63.2 & 55.6 & 68.0 & 72.0 & 61.8 & 69.28 \\
Swin-UNet~\cite{cao2022swinunet} & 97.4 & 56.0 & 96.1 & 84.7 & 71.0 & 59.3 & 53.9 & 73.8 & 73.9 & 71.09 \\
EGE-UNet~\cite{ruan2023ege} & 97.6 & 53.8 & 95.8 & 87.2 & 78.5 & 58.7 & 54.8 & 74.0 & 66.7 & 71.19 \\
FPN~\cite{lin2017fpn} & 97.7 & 53.8 & \underline{96.3} & 86.4 & 77.1 & \underline{60.2} & 62.3 & 74.4 & 66.9 & 72.17 \\
U-Net~\cite{ronneberger2015unet} & 97.8 & 55.7 & 96.1 & 87.4 & 71.0 & 58.1 & 67.7 & 75.2 & 69.6 & 72.62 \\
Mob.-UNETR~\cite{hatamizadeh2022unetr} & 97.5 & 56.7 & 95.9 & 80.9 & 73.9 & 59.7 & 68.9 & 70.5 & \underline{76.8} & 72.90 \\
SA-UNet~\cite{sun2022saunet} & \underline{97.9} & \textbf{57.3} & 94.4 & \underline{89.3} & 73.7 & 57.9 & \underline{71.9} & \textbf{78.7} & 70.5 & 74.21 \\
UNet3+~\cite{huang2020unet3p} & \underline{97.9} & 55.9 & \textbf{96.8} & 88.0 & \underline{79.4} & 58.7 & 66.2 & \underline{78.6} & 70.3 & \underline{74.22} \\
\midrule
\textbf{DeltaSeg} & \textbf{98.0} & \underline{56.5} & 95.9 & \textbf{89.6} & \textbf{80.5} & \textbf{60.3} & \textbf{76.7} & 77.5 & \textbf{76.9} & \textbf{76.75} \\
\bottomrule
\end{tabular}
}
\end{table}

\subsection{Ablation Study}

To isolate the contribution of the proposed attention and skip-connection mechanisms, we evaluate two reduced variants alongside the full model. DeltaSeg-V1 (1.96M) retains the DSC backbone with SE attention throughout and standalone Delta Operators on skips; DeltaSeg-V2 (5.44M) upgrades skips to full DDA modules while keeping SE attention; the full model (7.14M) additionally replaces decoder SE with Coordinate Attention and adds a CA layer at the bottleneck. \cref{tab:ablation} reports the results.

\begin{table}[!t]
\centering
\caption{Ablation study comparing DeltaSeg variants of increasing capacity on both datasets.}
\label{tab:ablation}
\resizebox{\columnwidth}{!}{%
\begin{tabular}{@{}lccc@{}}
\toprule
\textbf{Variant} & \textbf{Params (M)} & \textbf{\shortstack{S2DS \\ Def. mIoU (\%)}} & \textbf{\shortstack{CSDD \\ Def. mIoU (\%)}} \\
\midrule
DeltaSeg-V1 & 1.96 & 62.20 & 76.04 \\
DeltaSeg-V2 & 5.44 & 63.34 & 76.19 \\
\textbf{DeltaSeg (Full)} & \textbf{7.14} & \textbf{70.46} & \textbf{76.75} \\
\bottomrule
\end{tabular}
}
\end{table}

The V1$\to$V2 transition---replacing standalone Delta Operators with the full DDA module---yields +1.14 pp on S2DS and +0.15 pp on CSDD, attributable to DDA's decoder-conditioned spatial gating. The V2$\to$Full transition produces the largest gain (+7.12 pp on S2DS, +0.56 pp on CSDD), driven by Coordinate Attention's factored directional encoding that enables precise localization of elongated structures. The modest CSDD gain is consistent with its constrained cylindrical geometry. Even DeltaSeg-V1 (1.96M) outperforms most competing architectures, confirming that the DDA skip-connection design is the primary contributor to performance.

\begin{figure}[!t]
\centering
\includegraphics[width=\columnwidth]{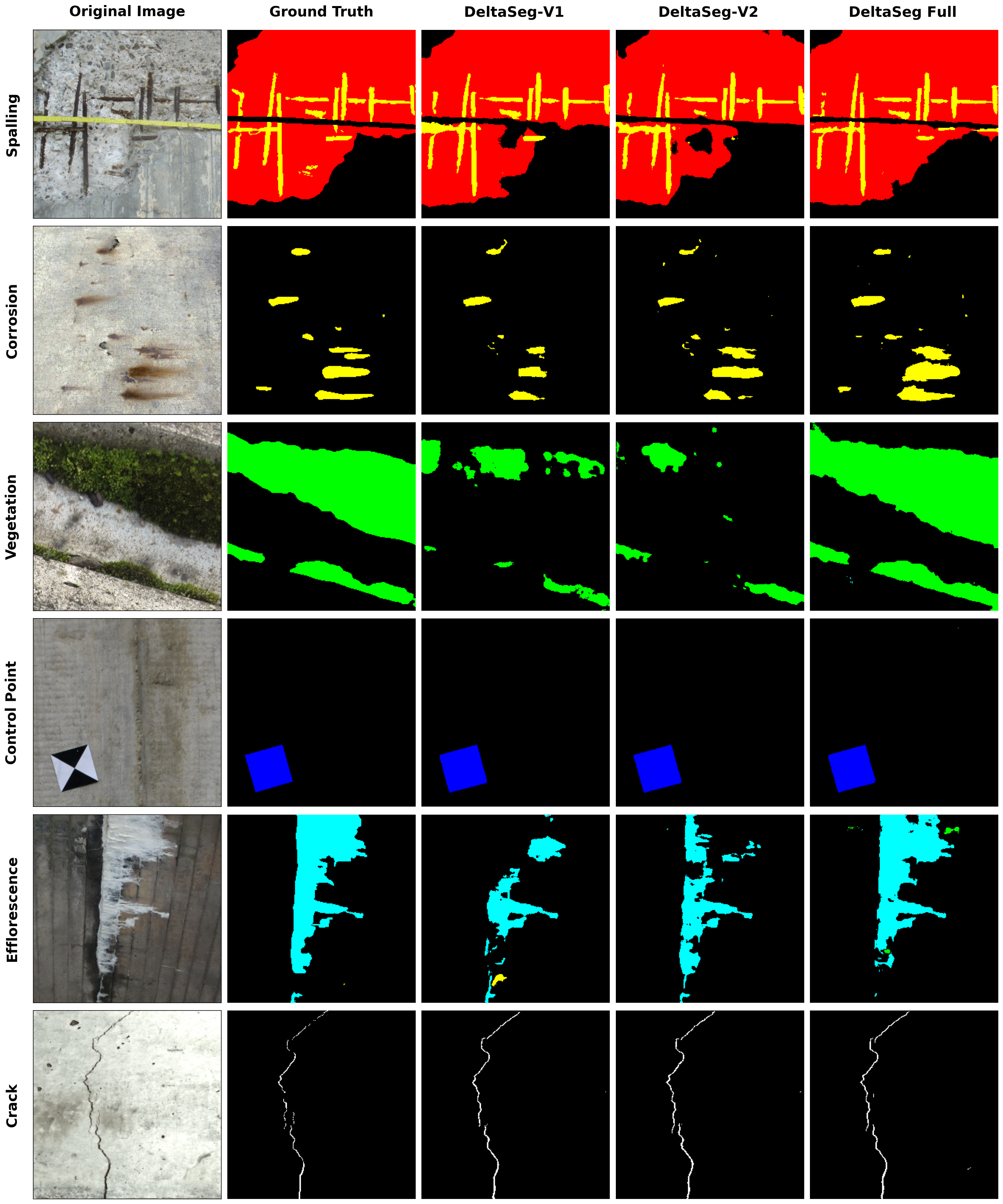}
\caption{Ablation comparison on S2DS. Rows: defect classes. Columns: image, ground truth, DeltaSeg-V1, V2, and Full model.}
\label{fig:ablation_comparison}
\end{figure}

\cref{fig:ablation_comparison} visualizes per-class segmentation quality across variants. The full model consistently outperforms V1 and V2 in boundary precision, with Coordinate Attention's directional encoding particularly evident on elongated defect boundaries.

\subsection{Computational Cost and Discussion}

DeltaSeg's depthwise separable backbone keeps the parameter count at 7.14M---substantially smaller than U-Net (31.04M), VM-UNet (29.08M), and UNet3+ (25.59M)---while achieving the highest accuracy on both datasets. The consistent advantage across different class counts (7 and 9), structural geometries (facades and culverts), and imaging conditions validates the tiered attention design: SE provides efficient channel calibration in the encoder, Coordinate Attention adds spatial awareness for boundary reconstruction in the decoder, and DDA addresses the semantic gap in skip connections.

\section{Conclusion}
\label{sec:conclusion}
This paper presented DeltaSeg, an encoder-decoder architecture with a tiered attention strategy for multi-class structural defect segmentation. By assigning SE attention to the encoder, Coordinate Attention to the bottleneck and decoder, and a novel Deep Delta Attention mechanism to the skip connections, the architecture matches each attention type to the specific requirements of its network location. The DDA module refines skip connections through a dual-path scheme that combines learned nuisance suppression with decoder-conditioned spatial gating. Evaluation on two datasets covering 7 and 9 classes, two distinct structure types, and diverse imaging conditions demonstrates consistent improvement over 12 competing architectures, including U-Net variants, attention-augmented models, transformer-based architectures, and lightweight designs. The results confirm that a tiered attention placement and refined skip connections provide both accuracy and generalization for structural defect segmentation.

Future work will investigate incorporating temporal information from sequential inspection images, extending the architecture to 3D point cloud segmentation, and deploying the model on edge devices for real-time field inspection.

\section*{Acknowledgments}
This research was supported by the U.S. Department of the Navy, Naval Research Laboratory (NRL) under contract N00173-20-2-C007. The views expressed in this paper are solely those of the authors and do not necessarily reflect the views of the funding agency.

{
    \small
    \bibliographystyle{ieeenat_fullname}
    \bibliography{references}
}

\end{document}